\documentclass[11pt,a4paper]{article}
\usepackage[hyperref]{emnlp2020}
\usepackage{times}
\usepackage{latexsym}
\usepackage{amsmath}
\usepackage{symbol}
\usepackage{multirow}
\usepackage{xspace,mfirstuc,tabulary}
\usepackage{booktabs}
\usepackage{slashbox}
\usepackage{xcolor,colortbl}

\usepackage{graphicx}  
\usepackage{microtype}

\usepackage{enumitem}

\usepackage{colortbl}

\usepackage{lipsum}

\usepackage{cleveref}
\crefformat{section}{\S#2#1#3}
\crefformat{subsection}{\S#2#1#3}
\crefformat{subsubsection}{\S#2#1#3}
\crefrangeformat{section}{\S\S#3#1#4 to~#5#2#6}
\crefmultiformat{section}{\S\S#2#1#3}{ and~#2#1#3}{, #2#1#3}{ and~#2#1#3}

\newcommand{\stitle}[1]{\vspace{0.3ex}\noindent{\bf #1}}

\aclfinalcopy 
\def\aclpaperid{189} 

\definecolor{ballblue}{rgb}{0.13, 0.67, 0.8}
\definecolor{azure1}{rgb}{0.0, 0.5, 1.0}
\definecolor{uclablue}{rgb}{0.33, 0.41, 0.58}
\definecolor{ultramarine}{rgb}{0.07, 0.04, 0.56}
\definecolor{yaleblue}{rgb}{0.06, 0.3, 0.57}

\usepackage{xcolor}

\newcommand{\modelname}{\texttt{P2GT}\xspace}

\newcommand{\ignore}[1]{}

\title{``What Are You Trying to Do?''\\
Semantic Typing of Event Processes}

\author{Muhao Chen$^{1,2}$, Hongming Zhang$^{3}$\thanks{\indent This work was done when the author was visiting the University of Pennsylvania.} , Haoyu Wang$^1$, Dan Roth$^1$\\
$^1$Department of Computer and Information Science, UPenn\\
$^2$Information Sciences Institute, USC\\
$^3$Department of Computer Science and Engineering, HKUST\\
\texttt{muhaoche@usc.edu};\;  \texttt{hzhangal@cse.ust.hk};\\ 
\texttt{\{why16gzi, danroth\}@seas.upenn.edu}
}

\date{2020/11}

\begin{document}
\maketitle

\begin{abstract}
This paper studies a new 
cognitively motivated semantic typing task, \emph{multi-axis event process typing}, that, given an event process, attempts to infer free-form type labels describing (i) the type of action made by the process and (ii) the type of object the process seeks to affect. 
This task is inspired by computational and cognitive studies of event understanding, which suggest that understanding processes of events is often directed by recognizing the goals, plans or intentions of the protagonist(s).
We develop a large dataset containing over 60k event processes, featuring ultra fine-grained typing on both the action and object type axes with very large ($10^3\sim 10^4$) label vocabularies.
We then propose a hybrid learning framework, \modelname, which addresses the challenging typing problem with indirect supervision from glosses\footnote{A gloss provides a sense definition for a lexeme.} and a joint learning-to-rank framework.
As our experiments indicate, \modelname supports identifying the intent of processes, as well as the fine semantic type of the affected object. It also demonstrates the capability of handling few-shot cases, and strong generalizability on out-of-domain processes.\footnote{The contributed learning resources, software and a system demonstration are available at \url{http://cogcomp.org/page/publication_view/915}.}
\end{abstract}



\section{Introduction}

Events are the fundamental building blocks of natural languages.
To help machines understand events, extensive research effort has been devoted to inducing how events described in text are procedurally connected \cite{ning-etal-2017-structured,radinsky2012learning}, and how they form \emph{event processes}\footnote{A.k.a. event chains \cite{chambers-jurafsky-2008-unsupervised}.} \cite{pichotta-mooney-2014-statistical,berant-etal-2014-modeling,JindalRo13d}.
Consequently, such prototypical schematic sequences of events have found important use cases including
storyline construction \cite{DoLuRo12,radinsky2013mining}, narrative cloze \cite{ChaturvediPeRo17,lee-goldwasser-2019-multi}, biological process comprehension \cite{berant-etal-2014-modeling} and diagnostic prediction \cite{zhang2020diagnostic}.

Nonetheless, understanding an event process is not just about inducing temporal relations between events or inferring missing steps in an event sequence. 
As suggested by cognitive studies~\cite{zacks2001human,zacks2001event,kurby2008segmentation}, a process of events is 
defined more by the goals, plans, intentions, or traits of its performer, rather than by physical characteristics.
For example, a series of events \emph{digging a hole}, \emph{putting in some seeds}, \emph{filling with soil} and \emph{watering the soil}, occurs in a 
specific sequence since these steps are directed towards the central goal of \emph{planting a plant} by the performer.  
Similarly, we can tell that \emph{making a dough}, \emph{adding toppings}, \emph{preheating the oven} and \emph{baking the dough} is likely a chain of actions aimed at \emph{cooking pizza}.
Indeed, aforementioned studies show that humans understand a plausible event process by hypothesizing the objectives those co-occurring events aim for, or the ultimate consequence the process seeks to accomplish.
Accordingly, we suggest that 
computational methods for event understanding would benefit from conceptualizing the intentions behind the processes.
Moreover, inducing intentions is crucial to rich understanding of text \cite{rashkin-etal-2018-event2mind}, and could potentially support other applications such as commonsense reasoning \cite{sap2019atomic}, summarization \cite{daume-iii-marcu-2006-bayesian}, reading comprehension \cite{berant-etal-2014-modeling} and schema induction \cite{huang-etal-2016-liberal}.

\begin{figure*}
    \centering
    \includegraphics[width=0.9\linewidth]{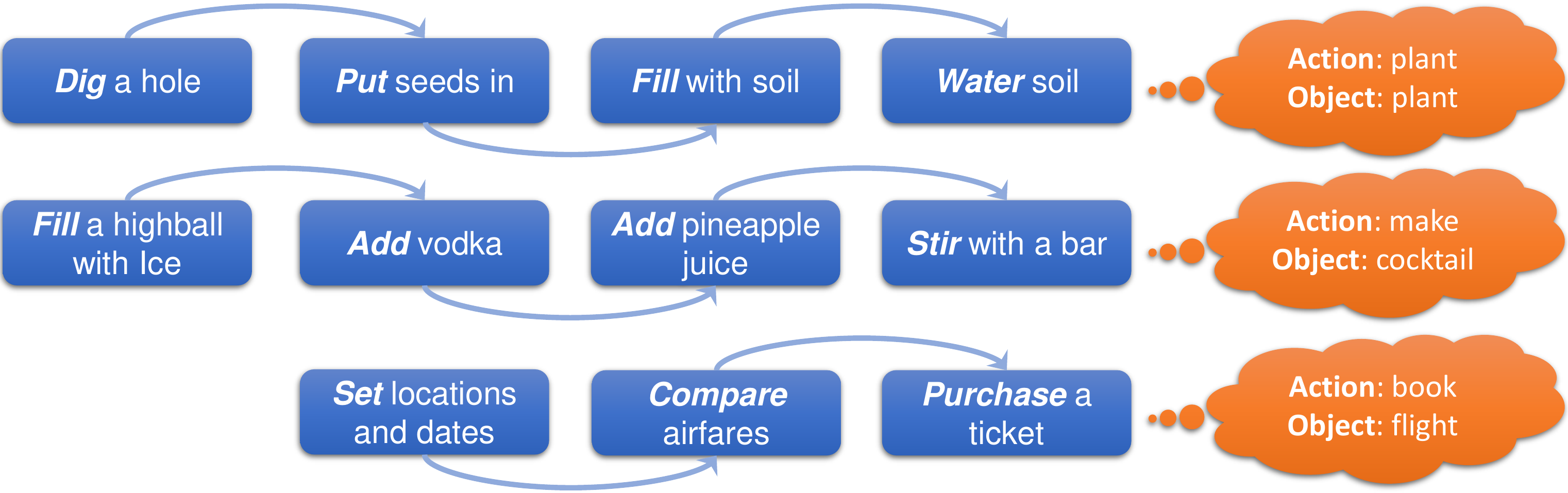}
    \caption{Examples of type inference for event processes.}\label{fig:example}
\end{figure*}

To understand the intentions of event processes, 
the \emph{first} contribution of this paper is to propose a new semantic typing task.
The 
\emph{event process typing} task seeks to retrieve ultra fine-grained type information to summarize the goal and intention of the associated events.
Specifically, each event process is typed along two axes: the \emph{action type} that describes the type of action the process takes, and the \emph{object type} that semantically types the object(s) that the process seeks to affect.
\Cref{fig:example} shows several accordingly typed event processes.
Motivated by 
recent works on entity typing
\cite{choi-etal-2018-ultra,zhou-etal-2018-zero},
our task employs large type vocabularies 
supporting diverse free-form semantic labels for both axes.  

To facilitate related research, we developed a large dataset extracted from wikiHow\footnote{\url{https://www.wikihow.com/}}, as the \emph{second} contribution of this paper.
This dataset contains over 60,000 
processes of primitive events, and features fine-grained action and object type labels for each process.
While the dataset aims at creating rich examples of event process intentions, it is also a challenging dataset from two perspectives.
First, vocabularies on both type axes are remarkably diverse, giving over 1,000
action type labels and over 10,000 
object type labels.
And, these fine-grained type vocabularies occur quite sparsely -- 
around 68\% of action types and 88\% of object types occur fewer than 10 times. This leads to a few-shot learning scenario and, in nearly half of the cases, one-shot.
Second, the free-form type labels are generally external to the lexical content of the associated events appearing in a process. 
Hence, this typing task could not be easily handled with an extractive method \cite{nenkova2012survey}.

While the task and dataset pose a non-trivial learning problem, 
the free-form type system allows for a practical form of indirect supervision based on gloss knowledge. 
As the \emph{third} contribution,
we propose a hybrid learning framework, \modelname (i.e., process-to-gloss based typing), to leverage such indirect supervision for event process typing.
Instead of directly inferring the multi-axis type labels, 
we find it to be much easier to seize on the semantic relatedness between the process-gloss pair, as the gloss provides richer semantic information than the label itself.
For few-shot cases, gloss definitions also represent useful side information to jump-start inducing labels that are rarely seen or completely unseen in training.

The proposed framework fine-tunes a pre-trained language model to capture the relatedness of an event process and the gloss of types with a ranking task objective.
To incorporate more precise gloss information, the training process deploys a word sense disambiguation (WSD) module for both verbs and nouns.
Joint learning for both action and object types is enforced to further complement scarce supervision signals.
Based on extensive experimental evaluation, the proposed framework exhibits promising performance of inferring the fine-grained multi-axis type information.
Specifically, it outperforms a strong RoBERTa-based baseline by 2.4-3.0 folds in \emph{recall@1}.
We also show that the incorporated gloss knowledge supports few-shot case prediction, and benefits our model's generalization to out-of-domain event processes.
\section{Task and Dataset}

We hereby formulate the 
task of multi-axis event process typing, and introduce the contributed dataset.

\subsection{Task Definition}
Following \citet{chambers-jurafsky-2008-unsupervised}, we define a process as a sequence of primitive events $P=[e_1,e_2,...,e_l]$ performed by one common protagonist (or performer).
Since the protagonist is shared among the events, each event $e_i$ thereof contains a \emph{predicate} $a_i$ mentioning an action performed by the protagonist, and an object $o_i$ describing the object(s) that the action is taken upon.
The goal is to conceptualize the overall intention behind the process $P$ into two labels, i.e. $A$ from the verb vocabulary that describes the overall action of $P$, and $O$ from the noun vocabulary that describes what object(s) the process is most likely to affect.
Such type inference is important to applications that require commonsense reasoning based on chains of activities, including event-based summarization, narrative prediction and open-domain QA.

\begin{figure}
    \centering
    \includegraphics[width=1.02\linewidth]{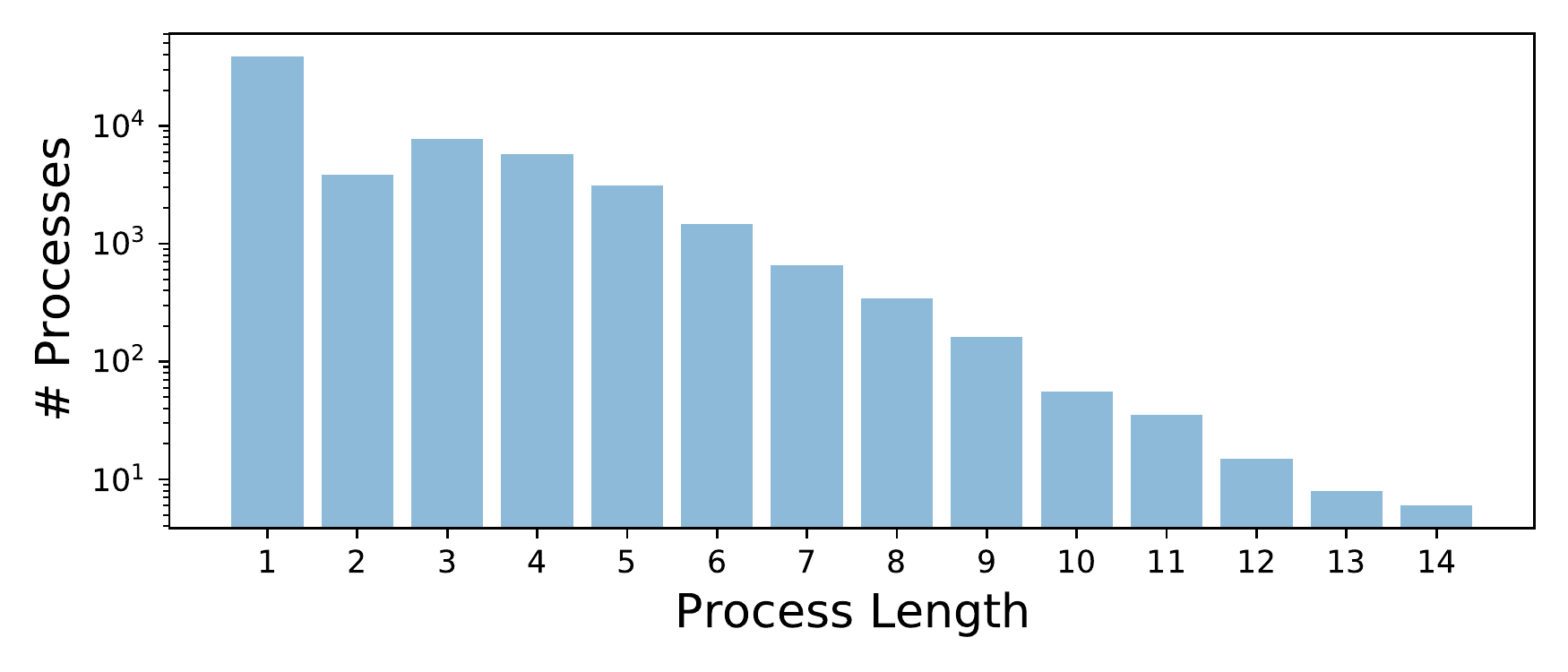}
    \vspace{-2em}
    \caption{Distribution of process lengths.}\label{fig:length_distribution}
\end{figure}

\begin{figure}
    \centering
    \includegraphics[width=1.0\linewidth]{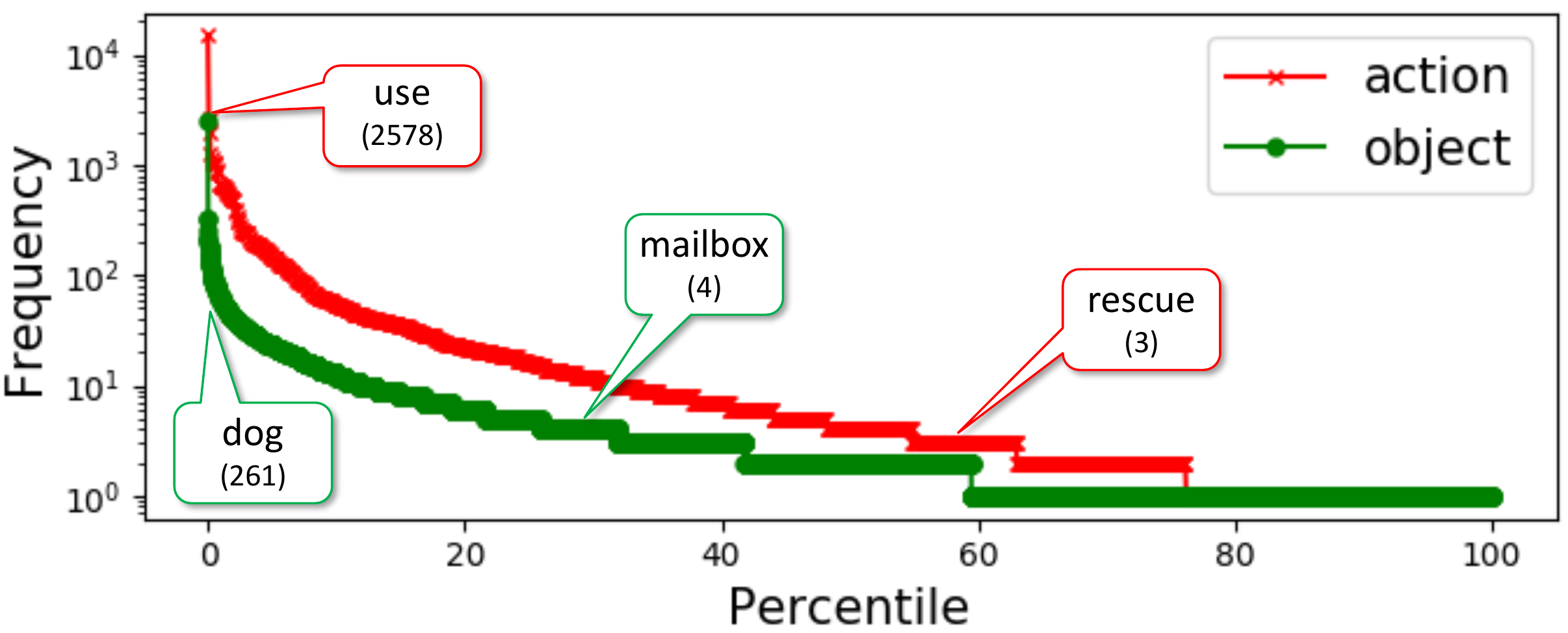}
    \caption{Distribution of action and object types. Number of frequencies are shown in the brackets.}\label{fig:distribution}
\end{figure}

\subsection{Dataset}
We construct a large corpus of typed event processes based on wikiHow -- an online wiki-style community containing a collection of professionally edited how-to guideline articles.

\paragraph{Construction} A set of the articles are crawled from wikiHow, where each included article describes ordered steps of activities to complete a central goal (e.g. the article ``How to book a flight'' describes necessary steps to complete an airline booking).
Each described step of an article forms a standalone section, which provides an easy-to-consume format for obtaining event processes with clear intentions.
We use AllenNLP \cite{gardner-etal-2018-allennlp} to perform SRL on section titles of a goal-step article, and extract the \texttt{VERB} (predicate $a_i$) and \texttt{ARG1} (object $o_i$) outputs from the section titles to form the corresponding sequence of (primitive) events.
Note that some articles may contain multiple step sequences for the same goal, e.g. booking a flight can be separated to two alternatives, either about booking online or via phone call. 
In such cases, each alternative is extracted as a separate process.
Moreover, we only preserve processes where every primitive event contains both \texttt{VERB} and \texttt{ARG1}.
Any \texttt{ARG0}'s are however omitted, since all events in a process share the same protagonist.

To obtain the type information, we first run SRL on the clause after ``how to'' in the article titles, from which the \texttt{VERB} term is seized as the action type label.
Then on the \texttt{ARG1} output of SRL, we fetch only the lemmatized head word based on dependency parsing and lemmatization \cite{bird-loper-2004-nltk}. This typically gives us the non-plural noun that represents the object type, whereas other dependents including modifiers are dropped.
Consider the clause in ``How to make a birthday cake'', after \emph{make} is fetched with SRL, the head word \emph{cake} will be preserved from the \texttt{ARG1} ``a birthday cake'',
providing an adequately abstracted label for object typing while being consistent to task definition.

\paragraph{Statistics}
The above effort obtains 62,277 clean event processes, each of which is labeled with both action and object types.
Lengths of the processes are varied, for which the distribution is plotted in \Cref{fig:length_distribution}.
While the dataset gives a rich variety of instances for processes and intentions,
it features a challenging type system for several reasons:

\begin{itemize}[leftmargin=1em]
\setlength\itemsep{0em}
\item \emph{Diversity.} The fine-grained type vocabularies consist of 1,336 action types and 10,441 object types. As shown in \Cref{fig:distribution}, both sets of labels generally form long-tail distributions. 
\item \emph{Few-shot cases.} There are 68.3\% of action type labels and 88.2\% of object type labels occuring fewer than 10 times across all processes. This fact indicates extreme few-shot cases that are challenging to learning and inference.
\item \emph{External labels.} In around 91.2\% processes, the action type labels are different from the predicates of associated events, while 84.2\% of processes have object type labels that do not appear as event objects. Such generally external labels 
easily cause extractive or sequence-to-label prediction methods to fall short.
\end{itemize}

\section{Process Typing with Gloss Knowledge}

In this section, we present our method for the multi-axis event process typing task.
The proposed \modelname framework conducts learning in three steps.
A pre-trained language model is first used to produce the representations of processes.
Then, the gloss information of type vocabularies is encoded as intermediate representations for type labels using the same language model, for which WSD is performed to refine the gloss information of polysemous labels during training.
Finally, the language model is fine-tuned with a ranking task objective to capture the association of process-gloss pairs.
In the last step thereof, joint learning is performed for typing on both axes to complement the scarce supervision signals, 
where a process representation is separately projected and handled for action and object types in the latent space.
Figure~\ref{fig:architecture} displays the overall model architecture.

In the rest of this section, we introduce the technical details of each step for learning and inference.

\subsection{Process Representation}\label{sec:3.1}

We use the officially released 
\emph{RoBERTa-base} \cite{liu2019roberta} for representations of event processes.
RoBERTa improves the original BERT \cite{devlin-etal-2019-bert} with a modified training procedure.
It is considered one of the SOTA models for semantic representation of lexical sequences.

To encode a process $P$, we concatenate the predicate and object ($a_i$ and $o_i$) of each event ($e_i)$.
Then those contents of all primitive events in $P$ are sequentially concatenated, while the separator token of RoBERTa \texttt{</s>} is added between the contents of every consecutive two events.
The entire lexical sequence is enclosed between tokens \texttt{<s>} and \texttt{</s>} to denote the beginning and end of the sequence.
Following convention \cite{bommasani-etal-2020-interpreting}, mean-pooling of hidden states produces the encoded representation of the process, denoted $\mathbf{P}$. 


\subsection{Label Representation}\label{sec:3.2}


In our problem setting, directly capturing the association between a process and a free-form type label can be difficult.
Hence, we propose a way of indirect supervision by using gloss knowledge as intermediate representations of type labels. 
The sense definitions in the glosses contain much richer semantic information of the labels themselves. Therefore, leveraging intermediate representations seeks to better characterize the semantic relatedness of processes and labels, especially when the labels are often external to the event content.
Glosses also adequately provide side information to jump-start few-shot label representations.

Given a label $L$ for either type axis,
we use the same RoBERTa model (with shared parameters) for process representation to encode its gloss sense definition.
The mean-pooling result produces the gloss-based label representation denoted $\mathbf{S}_L$.
Consider that the verb and noun terms in label vocabularies can be polysemous, we employ either of the following two techniques to select the glosses in the learning phase: 

\begin{figure}
    \centering
    \includegraphics[width=1.0\linewidth]{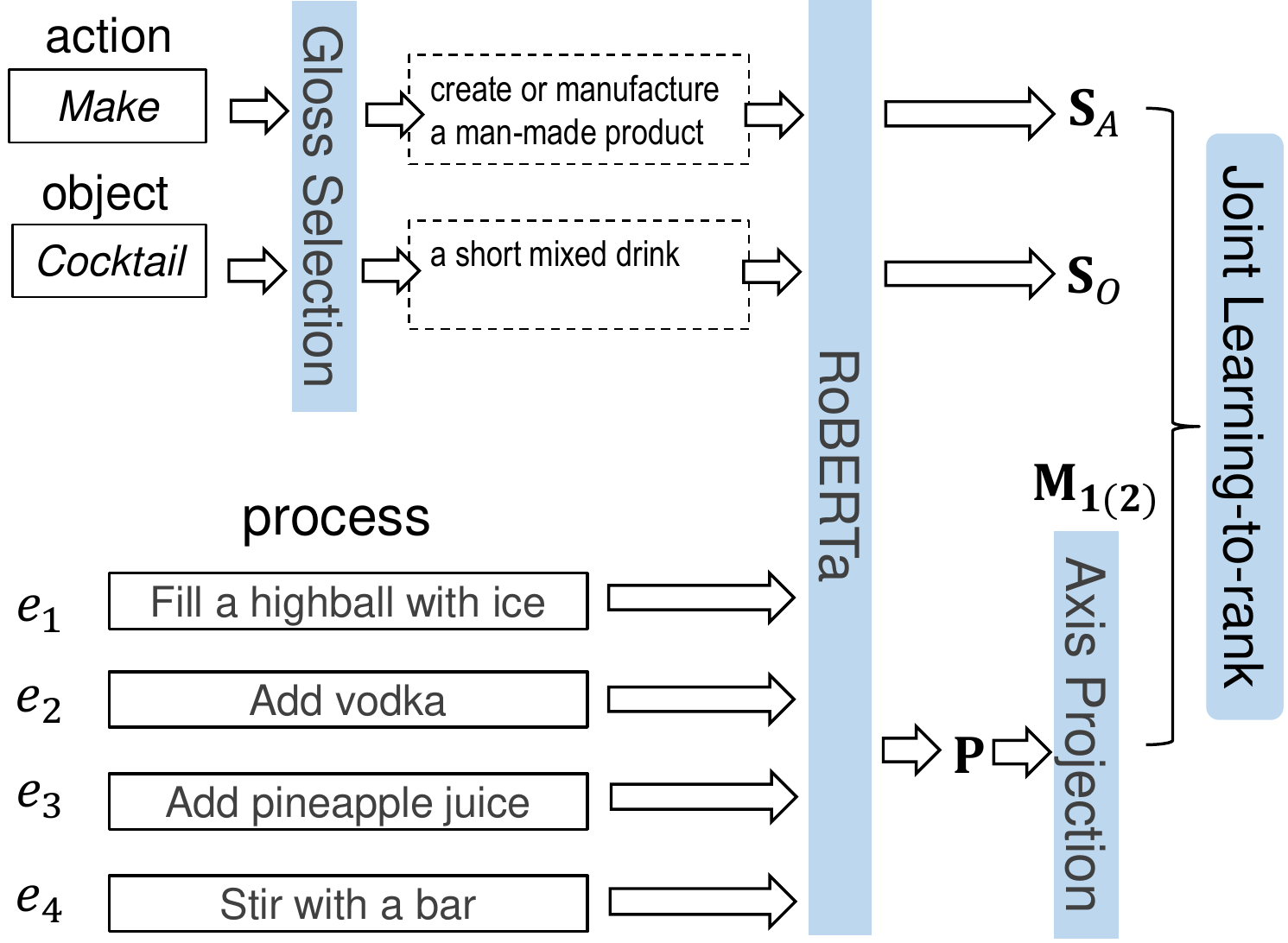}
    \caption{A gloss selection module selects the proper glosses of the training labels. Then a RoBERTa language model captures the event process, and separately generates gloss-based representations for positive and negative sampled labels. The entire learning process conducts joint learning-to-rank on both type axes.}\label{fig:architecture}
\end{figure}

\begin{itemize}[leftmargin=1em]
\setlength\itemsep{0em}
    \item \emph{Pre-trained WSD models}. One technique is to employ off-the-shelf WSD models that handle both verb and noun senses \cite{hadiwinoto-etal-2019-improved,huang-etal-2019-glossbert}. This could more precisely find the right definition for each label given the specific context of a process, and allows each (polysemous) label to have varied representations when typing different processes. During training \modelname, we run WSD on the concatenation of type labels $[A,O]$ to select the glosses of $A$ and $O$ for each training case. 
    \item \emph{Most frequent senses (MFS)}. Suppose a WSD model is not available, then the default way is to match a label only to its most frequent (or predominant) sense in sense-annotated corpora \cite{langone-etal-2004-annotating,camacho2016nasari}. The MFS method has been a very strong baseline for unsupervised WSD \cite{tripodi-navigli-2019-game}, as it is natural in language text that words generally express their predominant senses in most cases \cite{mccarthy-etal-2007-unsupervised}. Specifically for our task, the purpose is not to infer the exact sense, but rather generating a semantically rich (and allowably noisy) representation for type labels. In practice, we find this simple technique to perform reasonably well as we type the event processes (\Cref{sec:results}).
\end{itemize}

\noindent
Besides these two techniques, we also tried others to represent a label, including concatenating all its gloss sense definitions, or concatenating most frequent two or more senses.
They however do not perform as well as the aforementioned two techniques, hypothetically due to the noise introduced to label representations.
More technical details about WSD and the source inventory of glosses are to be described in Experiments (\Cref{sec:setting}).

\subsection{Learning Objective}\label{sec:3.3}

Let $(P,A,O)$ be a process $P$ denoted by action and object labels $A$ and $O$, our model captures the semantic associations between a RoBERTa encoded process $\mathbf{P}$ and label glosses $\mathbf{S}_A$ and $\mathbf{S}_O$ by optimizing a ranking task objective. 
In detail, we define the margin ranking loss for action typing as

\begin{equation*}
    L_1^P=\left [ s \left ( \mathbf{M}_1 \cdot \mathbf{P}, \mathbf{S}_{A'} \right ) - s \left ( \mathbf{M}_1 \cdot  \mathbf{P}, \mathbf{S}_{A} \right ) + \gamma_1 \right ]_+,
\end{equation*}

\noindent
and that for object typing as

\begin{equation*}
    L_2^P=\left [ s \left ( \mathbf{M}_2 \cdot \mathbf{P}, \mathbf{S}_{O'} \right ) - s \left ( \mathbf{M}_2 \cdot  \mathbf{P}, \mathbf{S}_{O} \right ) + \gamma_2 \right ]_+.
\end{equation*}
$[x]_+$ thereof denotes the positive part of the input $x$ (i.e. $\max(x, 0)$).
$\gamma_1$ and $\gamma_2$ are two positive constant margins. $\mathbf{M}_1$ and $\mathbf{M}_2$ correspond to two learnable linear projections dedicated to the two type axes respectively.
$s(\cdot)$ is the cosine similarity measure.
$A' \in V$ and $O' \in N$ are negative-sample labels.
In the setting with WSD deployed in training, negative sampling randomly fetches from all glosses of labels that appear in the training data, except for the gloss(es) of the positive label. This allows chances for different glosses of a polysemous label to serve as negative samples.
Otherwise, the only gloss of every negative-sample label is utilized in the MFS setting.

The eventual learning objective is to optimize the following joint loss, where $D$ denotes the dataset:
$$L = \frac{1}{|D|}\sum_{P\in D} L_1^P+L_2^P.$$
Note that we have incorporated different margins that can trade-off between $L_1^P$ and $L_2^P$, hence we do not use weight coefficients to combine these two terms of ranking losses.

\subsection{Inference}\label{sec:inference}

The inference phase of \modelname performs a nearest neighbor search to type a process $P$.
Let $\mathbf{M}$ refer to either $\mathbf{M_1}$ for the action type or $\mathbf{M_2}$ for the object type, our framework finds the gloss-based label representation that is closest to $\mathbf{M}\cdot \mathbf{P}$ from the corresponding vocabulary.
Specifically for the setting with polysemous label representations, it is sufficient to consider for each label only its gloss that is embedded most closely to $\mathbf{M}\cdot \mathbf{P}$,
so as to not redundantly consider candidate labels.

\section{Experiments}

To evaluate the proposed \modelname framework for event process typing,
we conduct several experiments on the contributed dataset, and compare with a wide selection of baseline methods (\Cref{sec:setting}-\Cref{sec:results}).
A case study is also provided on typing processes from an external dataset (\Cref{sec:case}).

\begin{table*}[!t]
    \centering
    {%
    \small
    \begin{tabular}{l||ccc|ccc}
    \midrule
    \toprule
     Type axes & \multicolumn{3}{c|}{Action} & \multicolumn{3}{c}{Object} \\ 
     \midrule
     Metrics & \emph{\;\; MRR\;\;} & \emph{recall@1} & \emph{recall@10} & \emph{\;\; MRR\;\;} & \emph{recall@1} & \emph{recall@10}\\
     \bottomrule
     S2L-mean-pool & 3.72 & 1.96 & 5.95 & 1.01 &0.80 & 1.66 \\
     S2L-BiGRU & 7.94 & 4.40 & 12.71 & 4.20 & 2.72 & 6.19 \\
     S2L-RoBERTa & 8.36 & 5.31 & 14.69 & 4.88 & 3.24 & 8.10 \\
     \midrule
     Single \modelname-MFS (partial event) & 18.03 & 14.36 & 17.16 & 10.36 & 6.37 & 17.64\\
     Single \modelname-WSD (partial event) & 18.07 & 14.05 & 17.82 & 10.72 & 6.68 & 18.03 \\
     Single \modelname-MFS & 24.10 & 19.67 & 32.40 & 13.71 & 8.86 & 23.09\\
     Single \modelname-WSD & 25.83 & 19.93 & 37.50 & 14.19 & 9.32 & 24.84\\
     \midrule
     Joint \modelname-MFS & 28.57 & 20.63 & \textbf{43.14} & 15.26 & 10.62 & 25.01\\
     Joint \modelname-WSD & \textbf{29.11} & \textbf{21.21} & 42.84 & \textbf{15.70} & \textbf{11.07} & \textbf{25.51}\\
    \bottomrule
    \end{tabular}
    }
    \vspace{-0.5em}
    \caption{Results (in percentage) for multi-axis event process typing. S2L methods with different encoding techniques are original or adopted from Event2Mind \cite{rashkin-etal-2018-event2mind}. \emph{partial event} marks the cases where only $a_i$ (or $o_i$) is encoded for each event $e_i$ in the process to infer the action (or object) type.  \emph{Joint} or \emph{Single} denotes whether to use joint training for both type axes or not. MFS and WSD marks ways of gloss selection in training.}
    \label{tbl:results}
\end{table*}
\begin{table*}[!t]
    \centering
    \setlength{\tabcolsep}{3pt}
    {%
    \small
    \begin{tabular}{p{0.52\linewidth}|p{0.45\linewidth}}
    \midrule
    \toprule
     \textbf{Event processes} & \textbf{Predictions} \\ 
     \midrule
     \multirow{2}{1.\linewidth}{Position yourself {\color{yaleblue}$\Rightarrow$} Trim your eyebrows {\color{yaleblue}$\Rightarrow$} Use the eyebrow pencil} & $A$: {\color{yaleblue}strop}, {\color{yaleblue}\textbf{highlight}}, \underline{\textbf{thread}}, {\color{yaleblue}blunt}, \emph{sharpen}\\
     & $O$: {\color{yaleblue}\textbf{unibrow}}, \underline{\textbf{eyebrow}}, {\color{yaleblue}straightener}, \textbf{eyelash}, {\color{yaleblue}razor}\\
     \midrule
     \multirow{2}{1.\linewidth}{Learn how to strum {\color{yaleblue}$\Rightarrow$} Use a metronome {\color{yaleblue}$\Rightarrow$} Play to recorded songs {\color{yaleblue}$\Rightarrow$} Grow skills} & $A$: \textbf{play}, \underline{\textbf{practice}}, {\color{yaleblue}\emph{strum}}, \emph{tune}, box\\
     & $O$: {\color{yaleblue}cymbal}, {\color{yaleblue}\textbf{mandolin}}, \underline{\textbf{guitar}}, {\color{yaleblue}\textbf{dulcimer}}, {\color{yaleblue}flute}\\
     \midrule
     \multirow{2}{1.\linewidth}{Get a referral {\color{yaleblue}$\Rightarrow$} Verify the specialist 's qualifications {\color{yaleblue}$\Rightarrow$} Ask questions {\color{yaleblue}$\Rightarrow$} Assess whether treatment is working} & $A$: \textbf{find}, \underline{\textbf{choose}}, use, apply, drink\\
     & $O$: {\color{yaleblue}\textbf{therapist}}, {\color{yaleblue}\emph{physician}}, {\color{yaleblue}\underline{\textbf{specialist}}}, {\color{yaleblue}\emph{surgeon}}, {\color{yaleblue}\emph{psychiatrist}}\\
     \midrule
     \multirow{2}{1.\linewidth}{Go to DMV {\color{yaleblue}$\Rightarrow$} Take photos {\color{yaleblue}$\Rightarrow$} Take vision test {\color{yaleblue}$\Rightarrow$} Take permit test {\color{yaleblue}$\Rightarrow$} Take road test} & $A$: \underline{\textbf{{obtain}}}, \emph{verify}, {\color{yaleblue}explore}, drive, polish\\
     & $O$: \underline{\textbf{{license}}}, check, {\color{yaleblue}visa}, {\color{yaleblue}carfax}, {\color{yaleblue}toll}\\
     \midrule
     \multirow{2}{1.\linewidth}{Create your clan {\color{yaleblue}$\Rightarrow$} Maintain your clan {\color{yaleblue}$\Rightarrow$} Add another clan {\color{yaleblue}$\Rightarrow$} Defend the borders {\color{yaleblue}$\Rightarrow$} Do the hunting} & $A$: \textbf{adopt}, \underline{\textbf{create}}, {\color{yaleblue}\textbf{spawn}}, {\color{yaleblue}\emph{homestead}}, {\color{yaleblue}become}\\
     & $O$: {\color{yaleblue}\textbf{clan}}, {\color{yaleblue}\underline{\textbf{{warrior}}}}, {\color{yaleblue}\emph{headhunter}}, {\color{yaleblue}skirmish}, {\color{yaleblue}\emph{necrons}}\\
     \midrule
     \multirow{2}{1.\linewidth}{Prepare the jack {\color{yaleblue}$\Rightarrow$} Locate the filler hole {\color{yaleblue}$\Rightarrow$} Fill the oil {\color{yaleblue}$\Rightarrow$} Close the filler hole} & $A$: {\color{yaleblue}\emph{bleed}}, {\color{yaleblue}\emph{grease}}, \underline{\textbf{add}}, \textbf{fill}, inflate\\
     & $O$: \underline{\textbf{{oil}}}, {\color{yaleblue}pump}, {\color{yaleblue}\emph{biodiesel}}, {\color{yaleblue}blowing}, {\color{yaleblue}choke}\\
    \bottomrule
    \end{tabular}
    }
    \vspace{-0.5em}
    \caption{Top 5 predictions on examples of test cases by Joint-\modelname-WSD. Ground truths are underscored, reasonably correct labels are boldfaced, and close ones are italic.  Few-shot labels appearing $\leq$10 times are in blue.}\label{tbl:predict}
\end{table*}

\subsection{Experimental Settings}\label{sec:setting}

Similar to \citet{rashkin-etal-2018-event2mind}, we randomly separate the 62,277 processes into a training/dev./test set using an 80/10/10\% split.
We report three ranking metrics, i.e. \emph{MRR} (mean reciprocal rank), \emph{recall@1} and \emph{recall@10}. All metrics are preferred to be higher to indicate better performance.

We compare our framework with a number of its variants by performing the following modification:
(i) Simplifying the framework by separately learning for the two type axes, instead of performing joint training;
(ii) Different settings of gloss selection in training, using either WSD or FSM;
(iii) Different information used to represent each primitive event $e_i$, e.g., only using either $a_i$ or $o_i$ (marked with \emph{partial event}) according to the type axis, instead of using both.
Besides, we compare with sequence-to-label (S2L) generators \cite{rashkin-etal-2018-event2mind}.
A method of such is an encoder-decoder architecture trained to directly map from sequences to unigrams of the type vocabulary, which is originally used by recent work \cite{rashkin-etal-2018-event2mind} to infer intentions from a single-clause description of a primitive event.
Specifically, we employ three variants of S2L using different encoders. Besides one based on RoBERTa (marked as S2L-RoBERTa), the two others are the BiGRU encoder (S2L-BiGRU) and mean-pooling encoder (S2L-mean) with Skip-Gram word embeddings used by \citet{rashkin-etal-2018-event2mind}. Note that to train S2L models, the original paper uses an cross-entropy loss to model the distribution of unigrams. We instead train the process encoder to directly fit the embeddings of label surface forms similar to a reverse dictionary \cite{hill2016dict,chen-etal-2019-bildict}, which offers notably better performance.



\subsection{Model Configuration}

We use sense definitions from WordNet \cite{miller1995wordnet} to define the labels.
While such glosses cover all verbs in the action type vocabularies, there are 7.92\% of processes where object type labels do not find WordNet senses.
For each such case, we select from WordNet the lexeme that is embedded most closely to the label, and use the predominant sense of that lexeme to generate the label representation.
For the training setting with WSD, we use the BERT-NN model \cite{hadiwinoto-etal-2019-improved}, which is one of the SOTA WSD methods that is trained on the SemCor corpus \cite{langone-etal-2004-annotating}.
In fact, despite the ones that are dedicated to nouns \cite{scarlini2020sensembert,pasini-navigli-2017-train}, other SOTA methods for WSD \cite{huang-etal-2019-glossbert,maru-etal-2019-syntagnet,tripodi-navigli-2019-game} may also apply to our framework, for which we leave the investigation to future work.

We use AMSGrad \cite{reddi2018convergence} to optimize the learning objective, with the learning rate set to 0.0001. The batch size is set to 64 to fit the memory of one Titan RTX 6000 GPU. Training is limited to 50 epochs that is enough for all models to converge. Margins are chosen from 0.0 to 0.4 with a step of 0.1, based on \emph{recall@1} performance on the dev. set. Accordingly, $\gamma_1=0.2$ and $\gamma_2=0.1$ are selected for Single \modelname methods, while both margins are set to $0.1$ for the joint-learning \modelname.

\begin{figure}[t]
    \centering
    \includegraphics[width=1.0\linewidth]{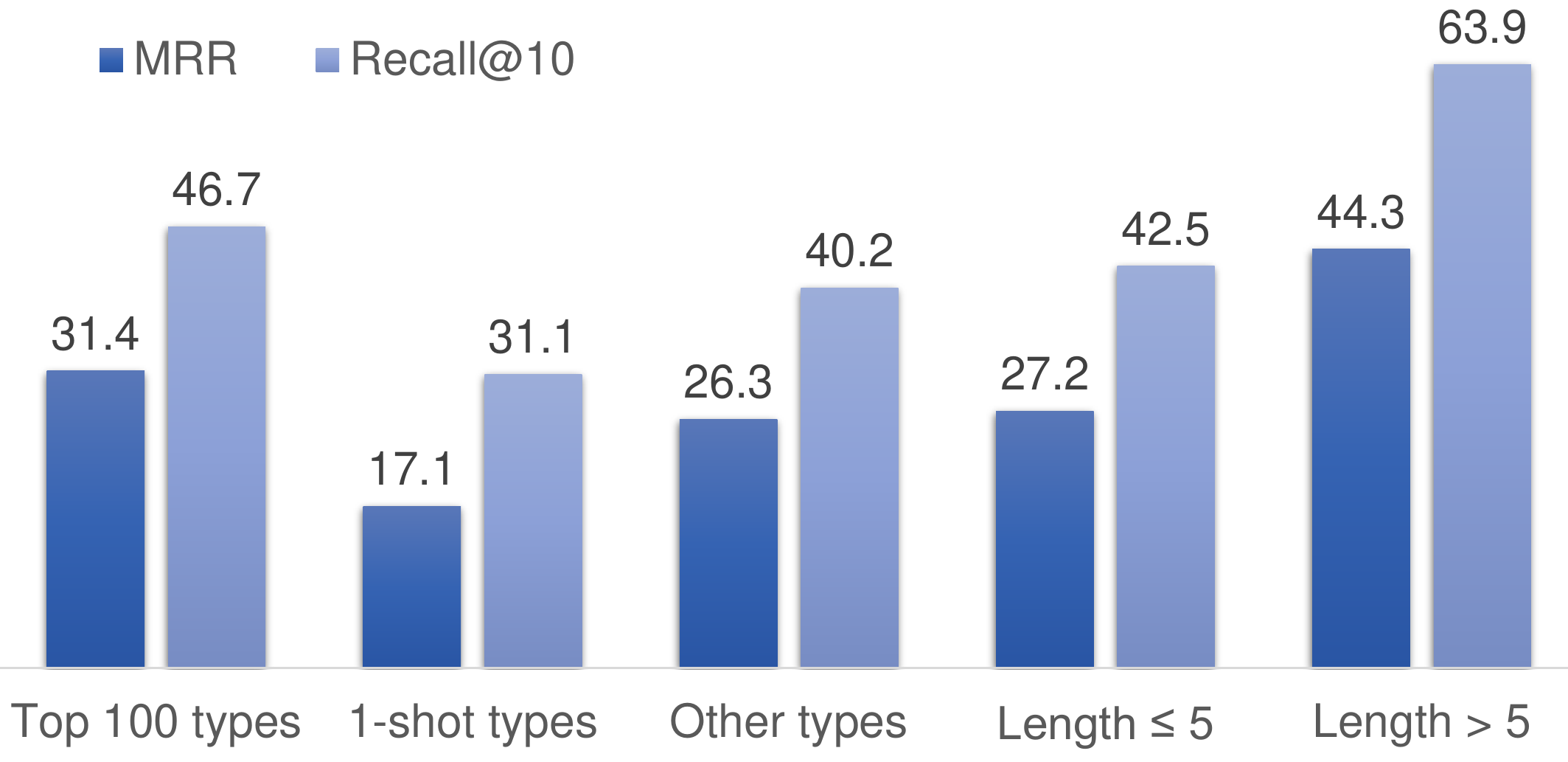}
    \caption{Comparison of action typing on different portions of the test set. We compare results by Joint \modelname-MFS for top 100 frequent types, one-shot types and the rest, as well as results on processes of different lengths. 5 is the median length of processes in the dataset.}
    \label{fig:group_results}
\end{figure}

\begin{table*}[!t]
    \centering
    \setlength{\tabcolsep}{3pt}
    {%
    \small
    \begin{tabular}{p{0.66\linewidth}|p{0.3\linewidth}}
    \midrule
    \toprule
     \textbf{Event processes} & \textbf{Predictions} \\ 
     \midrule
     \multirow{2}{1.\linewidth}{Make explosive materials {\color{yaleblue}$\Rightarrow$} Obtain a container {\color{yaleblue}$\Rightarrow$} Obtain shrapnel {\color{yaleblue}$\Rightarrow$} Install a trigger} & $A$: {\color{yaleblue}\textbf{{detonate}}}, \textbf{{assemble}}, {\color{yaleblue}blacken}\\
     & $O$: \textbf{{grenade}}, \textbf{{blaster}}, {\color{yaleblue}\textbf{{mine}}}\\
     \midrule
     \multirow{2}{1.\linewidth}{Ignore order {\color{yaleblue}$\Rightarrow$} Enter area {\color{yaleblue}$\Rightarrow$} Enforce blockade {\color{yaleblue}$\Rightarrow$} Force to retreat from area} & $A$: {\color{yaleblue}\textbf{{conquer}}}, {\color{yaleblue}\emph{disarm}}, {\color{yaleblue}\textbf{{invade}}}\\
     & $O$: {\color{yaleblue}\emph{barrier}}, {\color{yaleblue}\textbf{{soldier}}}, {\color{yaleblue}\textbf{{fortress}}}\\
     \midrule
     \multirow{2}{1.\linewidth}{Capture two opposition posts {\color{yaleblue}$\Rightarrow$} Kill many fighters {\color{yaleblue}$\Rightarrow$} Destroy three armed trucks {\color{yaleblue}$\Rightarrow$} Confiscate artillery guns} & $A$: \emph{kill}, {\color{yaleblue}\textbf{demolish}}, \textbf{{fight}}\\
     & $O$: \emph{melee}, {\color{yaleblue}\textbf{{conflict}}}, {\color{yaleblue}\textbf{{stronghold}}}\\
     \midrule
     \multirow{2}{1.\linewidth}{Cooperate with the counsel investigation {\color{yaleblue}$\Rightarrow$} Open his remarks {\color{yaleblue}$\Rightarrow$} Apologize many times {\color{yaleblue}$\Rightarrow$} Try to restore public trust} & $A$: \emph{respond}, {\color{yaleblue}disagree}, \textbf{{accept}}\\
     & $O$: {\color{yaleblue}\emph{apology}}, {\color{yaleblue}\emph{disagreement}}, {\color{yaleblue}\textbf{{slander}}}\\
     \midrule
     \multirow{2}{1.\linewidth}{Travel in a presidential motorcade {\color{yaleblue}$\Rightarrow$} Be shot once in the back {\color{yaleblue}$\Rightarrow$} Be taken to hospital {\color{yaleblue}$\Rightarrow$} Be pronounced dead} & $A$: \emph{survive}, {\color{yaleblue}\textbf{{die}}}, {\color{yaleblue}tackle}\\
     & $O$: {\color{yaleblue}\textbf{assassin}}, {\color{yaleblue}crash}, {\color{yaleblue}\emph{{roadkill}}}\\
     \midrule
     \multirow{2}{1.\linewidth}{Give advance notice {\color{yaleblue}$\Rightarrow$} Give notice {\color{yaleblue}$\Rightarrow$} Issue dividends} & $A$: {\color{yaleblue}\textbf{{honor}}}, \textbf{{pay}}, {\color{yaleblue}\textbf{{reward}}}\\
     & $O$: {\color{yaleblue}\emph{finance}}, {\color{yaleblue}\textbf{{equity}}}, {\color{yaleblue}\textbf{{subsidy}}}\\
     \midrule
     \multirow{2}{1.\linewidth}{Target quotes {\color{yaleblue}$\Rightarrow$} Target shares quotes {\color{yaleblue}$\Rightarrow$} Ask to clarify offer {\color{yaleblue}$\Rightarrow$} Challenge to merge agreement {\color{yaleblue}$\Rightarrow$} Challenge to merge businesses} & $A$: \textbf{{compare}}, \textbf{{maximize}}, {\color{yaleblue}\textbf{{negotiate}}}\\
     & $O$: {\color{yaleblue}\emph{prospectus}}, \textbf{{quote}}, {\color{yaleblue}\textbf{{settlement}}}\\
     \midrule
     \multirow{2}{1.\linewidth}{Clean windows {\color{yaleblue}$\Rightarrow$} Buy plants {\color{yaleblue}$\Rightarrow$} Hang pictures {\color{yaleblue}$\Rightarrow$} Paint walls {\color{yaleblue}$\Rightarrow$} Carpet floors} & $A$: \textbf{{\color{yaleblue}{redecorate}}}, \textbf{{decorate}}, {\color{yaleblue}\emph{refurbish}}\\
     & $O$: \textbf{{room}}, \textbf{{bedroom}}, \emph{makeover}\\
    \bottomrule
    \end{tabular}
    }
    \caption{Case study for typing event processes in the news domain. The predictions are given by Joint \modelname-WSD trained on our full dataset. Each case is given top 3 predictions on both axes, whereof reasonably correct ones are boldfaced, and relevant ones are italic. Few-shot labels appearing up to 10 times in our dataset are in blue.}
    \label{tbl:case}
\end{table*}

\subsection{Results}\label{sec:results}

We report the results of event process typing on both axes in \Cref{tbl:results}, whereof the results for typing actions are generally better than those for the object axis due to different sizes of candidate spaces.

The results by the S2L baseline methods show that incorporating pre-trained RoBERTa offers noticeably better performance than other encoding techniques. However, it is drastically superseded by the Single \modelname setting without WSD-based gloss selection. 
When typing the action, with the same representation of event processes, \modelname supercedes S2L-RoBERTa with an absolute increase of \emph{MRR} by 15.47\% (ca. 1.88$\times$ relative increase), and that of \emph{recall@1} by 14.36\% (ca. 2.7$\times$ relative increase). 
For object typing, the absolute increments are 8.83\% in \emph{MRR} (ca. 1.82$\times$ relatively) and 5.62\% in \emph{recall@1} (ca. 1.72$\times$ relatively). 
This also indicates that incorporating gloss knowledge for label representations brings along the most substantial improvement to the task,
inasmuch as glosses attain rich semantic information to jump-start type labels and realistically support sequence-level learning-to-rank.

On the other hand, incorporating WSD for gloss selection in training slightly causes absolute increment of up to 1.73\% in \emph{MRR} for action typing and 0.48\% in \emph{MRR} for object typing.
This is partly due to that the predominant sense definitions can generally seize precise or close definitions to represent the labels in most cases.
Hence, sense selection provides lesser improvement, especially when the candidate space is large.
It is noteworthy that, partially giving the predicate or object information of associated events is not enough to infer the type information. 
In fact, the performance drop is in accordance with human cognition, as giving a chain of predicates only or objects only is not enough to predict the intentions of the event process.
Consider the first example in \Cref{fig:example}, observing only a chain of the protagonist's actions \emph{dig}, \emph{put}, \emph{fill} and \emph{water}, or only the participating objects \emph{hole}, \emph{seed} and \emph{soil} are clearly not enough to infer the overall action and the objective that directs the entire process.
Accordingly, the partial event representation causes significant performance drop of 3.35-7.76\% in terms of \emph{MRR}, and 2.49-5.86\% in terms of \emph{recall@1}. 
Lastly, joint learning brings along performance gain by 1.51-4.47\% in \emph{MRR} and 0.96-1.76\% in \emph{recall@1}, indicating the effectiveness of leveraging complementary supervision signals.
Note that the evaluation strictly enforces \emph{exact match} in large candidate spaces, thus underestimating the system performance. While it is difficult for the model to always rank the ground-truth labels on the top, it can often infer reasonably close labels as top predictions, for which a couple of examples are shown in \Cref{tbl:predict}.

To understand how differently our method performs on processes of different characteristics, we additionally perform an error analysis.
In \Cref{fig:group_results}, we compare action prediction by \modelname with joint learning and MFS-based gloss selection on different proportions of the test set. 
It is expected that the performance on more frequent labels are better than on infrequent ones due to ampler training cases. Nonetheless, on the extremely challenging one-shot cases, our method still performs reasonably well, and drastically excels the overall results by baseline methods.
Additionally, we observe that typing longer event processes is easier, as they provide more contextual information of associated events to help inferring the central goal.
In contrast, as short processes are less informative, \emph{MRR} scores for those sized 2 and 3 are 24.17\% and 25.41\%.


\subsection{Case Study}\label{sec:case}

We conduct a case study using a subset of the NYT narrative cloze dataset provided by \citet{lee-goldwasser-2019-multi}. 
This dataset includes a series of event processes extracted from news reports, and we use those processes to showcase the prediction of \modelname on out-of-domain processes.
According to \Cref{tbl:case}, although the content and concepts of processes in military and political news are mostly irrelevant to the intentional goals in our dataset, \modelname is able to infer reasonably correct type information on both axes.
Particularly, many of the top predictions give few-shot labels.
This further exhibits that gloss knowledge is effective to improve the generalization of the typing model, both in terms of handling domain shifting and few-shot cases. Specifically, the case study also points out the direction of our further study on how well gloss-based label representations can generally benefit domain adaptation and few-shot learning in natural language understanding tasks.
\section{Related Work}


\stitle{Prediction tasks on event processes} have attracted much attention recently,
while many existing works focus on extraction and completion of event processes.
For example, \citeauthor{radinsky2012learning}~\shortcite{radinsky2012learning,radinsky2013mining} mine sequences of frequently co-ocurring events from multiple temporally connected documents, and use the sequence knowledge to predict the future event(s) of a process.
\citet{berant-etal-2014-modeling} propose to extract biological processes with SRL, and help machine reading comprehension for biological articles.
A series of other works learn for sequential event prediction using language models \cite{ChaturvediPeRo17,peng-etal-2019-knowsemlm} or association rules \cite{letham2013sequential}, and further cope with downstream tasks such as narrative cloze tests.
On the contrary, fewer efforts have been made for inferring the intentions or central goals behind a composite of events.
A recent work by \citet{rashkin-etal-2018-event2mind} is particularly relevant to this topic, which learns a sequence-to-label generator to predict the intention of one primitive event based on a single-clause description. This is however essentially different from our focus on processes of multiple events.

\stitle{Semantic typing} has been investigated for language components other than events, such as entities and word senses.
Due to the large body of work in this line of research, we can only provide a highly selected summary for most recent outcomes.
For entity typing, recent research has coped with highly challenging problem settings.
Those include few-shot or zero-shot typing with contextual distant supervision \cite{zhou-etal-2018-zero} and description-based label embeddings \cite{obeidat-etal-2019-description}.
Others realize ultra-fine type systems with the help of head-word supervision \cite{choi-etal-2018-ultra}, hierarchical learning-to-rank \cite{chen-etal-2020-hierarchical} and structured label representations \cite{xiong-etal-2019-imposing,hao2019universal}.
Several aforementioned techniques are also employed to supersense typing \cite{levine-etal-2020-sensebert,peters-etal-2019-knowledge} and POS tagging \cite{owoputi-etal-2013-improved}.
In terms of type labeling, our work is inspired by \citet{choi-etal-2018-ultra}'s way of leveraging free-form lexemes for ultra-fine entity types.
Nevertheless, besides typing on a different modality, our work is also distinguished in the multi-axis typing system, and the way of leveraging gloss-based indirect supervision.

\stitle{Representation learning of gloss knowledge} has been incorporated in various tasks.
A number of works encode gloss definitions for monolingual \cite{hill2016dict,noraset2017definition,pilehvar-2019-importance,hedderich-etal-2019-using} and cross-lingual \cite{chen-etal-2019-bildict,zhang2020multi} reverse dictionary prediction, as well as out-of-vocabulary lexical representation
\cite{kumar-etal-2019-zero,prokhorov2019unseen,bahdanau2017fly}.
Definitions have also been leveraged to generate zero-shot entity representations in knowledge graphs \cite{kartsaklis-etal-2018-mapping,chen2018co,long-etal-2017-world}.
Some other works inject gloss representations to improve WSD \cite{huang-etal-2019-glossbert,luo-etal-2018-leveraging,blevins-zettlemoyer-2020-moving}.
GlossBERT \cite{huang-etal-2019-glossbert} thereof formalizes the WSD problem as classifying context-gloss pairs.
Our learning approach on process-gloss pairs is connected to that approach, whereas we handle a learning-to-rank objective, and make inference in a much larger candidate space than the sense space of a single word.

\section{Conclusion}

We propose a new task of event process understanding, by 
semantically typing the intended action of an event process 
and the
object(s) it seeks to affect.
To facilitate 
research in this direction, we develop
a new dataset, gathering over 60 thousand event processes
with ultra fine-grained type vocabularies.
We further propose a hybrid learning framework, which leverages indirect supervision from gloss knowledge.
The proposed \modelname framework fine-tunes RoBERTa to capture the association of process-gloss pairs.
Label gloss selection mechanisms and joint training are incorporated to further improve the performance.
Experiments show that \modelname offers promising performance on inferring the fine-grained type information, and exhibits satisfactory generalizability on out-of-domain event processes. 

For future work, we are interested in identifying salient events in processes, i.e., those that most significantly define the central goals. Incorporating process typing into downstream tasks such as summarization and commonsense QA is also an important
direction.
\section*{Acknowledgement}

We appreciate the anonymous reviewers for their insightful comments.
Also, we would like thank Jennifer Sheffield and other members of
the UPenn Cognitive Computation Group for giving suggestions 
that improved 
the manuscript.


This research is supported by the Oﬃce of the Director of National Intelligence (ODNI), Intelligence Advanced Research Projects Activity (IARPA), via IARPA Contract No. 2019-19051600006 under the BETTER Program, and by Contract FA8750-19-2-1004 with the US Defense Advanced Research Projects Agency (DARPA). The views expressed are those of the authors and do not reflect the official policy or position of the Department of Defense or the U.S. Government. 

\bibliography{anthology,emnlp2020,ccg.bib}
\bibliographystyle{acl_natbib}

\end{document}